\documentclass[lettersize,journal]{IEEEtran}
\usepackage{amsmath,amsfonts,amssymb}
\usepackage{algorithmic}
\usepackage{algorithm}
\usepackage{array}
\usepackage[caption=false,font=normalsize,labelfont=sf,textfont=sf]{subfig}
\usepackage{textcomp}
\usepackage{stfloats}
\usepackage{url}
\usepackage{graphicx}
\usepackage{cite}
\usepackage{multirow}
\usepackage{booktabs}
\usepackage{bm}
\usepackage{hyperref}
\newcommand{\sdf}{\operatorname{sdf}}
\newcommand{\relu}{\operatorname{relu}}
\newcommand{\sg}{\operatorname{sg}}

\begin{document}

\title{\fontsize{22pt}{30pt}\selectfont Some Modifications to Our End-to-End UAV Planner}
\author{Junjie Lu, Bailing Tian*}

\maketitle

\begin{abstract}
The one-stage planner YOPO maps a single depth image and the robot state directly to a set of candidate trajectories, trained by backpropagating through differentiable trajectory costs. This yields dense, geometrically informative supervision, but inherits the pathologies of soft-constrained optimization: the safety cost competes with the smoothness and goal-reaching terms, is non-convex across homotopy classes, and the single-piece polynomial is limited in expressiveness.

In this report, we summarize several effective modifications. We adopt a two-piece MINCO parameterization, trading time for smoothness without altering the trajectory's spatial profile. We further lift YOPO's multi-modal prediction to span distinct homotopy classes, treating each motion primitive as a homotopy anchor that confines the trajectory to a feasible basin—without explicit safe-flight-corridor construction or front-end search. For dynamic feasibility, we impose barrier penalties on velocity and acceleration together with a curvature-dependent speed limit whose gradient acts only on the velocity, producing an adaptive-speed behavior that decelerates in cluttered regions or sharp turns. We replace score regression with a ranking loss, preventing small score errors from reordering the candidate set. These yield richer trajectory representations, safer obstacle avoidance, and more direct flight paths.
\textbf{Code:} \url{https://github.com/TJU-Aerial-Robotics/YOPO/tree/YOPO-MINCO}.
\end{abstract}


\section{Introduction}

\IEEEPARstart{L}{arge} models have advanced so rapidly in recent years, making robots increasingly intelligent. Before shifting my research direction, I have summarized the useful attempts to improve YOPO \cite{ref_yopo} below to document the development of the open-source branch.

Classical planners \cite{fast} decompose navigation into mapping, front-end path search, and back-end optimization, which has evolved into a mature and well-established pipeline. The learning-based planners avoid explicit mapping and are inherently data-driven, achieving impressive performance in high-speed flight with only a depth camera and a resource-constrained onboard computer. Among these, some methods fit the trajectory with a single-piece fifth-order polynomial \cite{SR} or a spline \cite{Deep-panther}, while more of them let the network predict control commands directly \cite{back, racing}. Predicted trajectories are relatively smooth and have been studied extensively in classical methods, whereas predicting control commands achieves more aggressive performance.

In \cite{ref_yopo}, we leverage the multi-modal commonality shared by navigation and detection tasks to enable multi-modal prediction through a set of predefined primitive anchors. Instead of relying on expert actions or scalar environment rewards, the network is trained with differentiable trajectory costs borrowed from classical trajectory optimization. This yields dense, geometrically accurate supervision without online rollout or rendering.

A limitation is intrinsic to this paradigm: this supervision is essentially a weighted sum of multiple soft constraints, and different weight combinations steer the optimizer toward different local optima. This is a shared drawback of the soft-constraint paradigm itself, rather than a problem specific to learning-based methods: imitation learning requires tuning the weights within the expert policy that generates the demonstrations, while reinforcement learning requires tuning the weights within the reward function.
Classical methods typically rely on front-end search to identify a local minimum in advance for soft-constrained gradient optimization, or place waypoints at well-chosen turning points—often the interfaces between adjacent corridors—to solve a hard-constrained problem. 

Hard-constrained methods in particular have achieved impressive high-speed flight, although the corridor is sensitive to noise, which confines them mostly to high-accuracy LiDAR. They first extract a collision-free corridor and then optimize within it, thereby anchoring the solution to a homotopy class, while the corridor junctions provide well-placed turning points for initialization. Bernstein-basis
formulations~\cite{ref_bezier} exploit the convex-hull property of B\'ezier curves to keep
the whole trajectory inside the corridor---elegant, but conservative. MINCO~\cite{ref_minco}
relaxes this by constraining only the waypoints to lie on the interfaces between
consecutive corridors through a diffeomorphic transformation, while the trajectory segments
themselves are handled softly, demonstrating remarkable performance. Notably, several MINCO-based works~\cite{ref_bubble} achieve impressive
results without enforcing hard constraints at all, penalizing only the distance from
each trajectory sample to the center of a spherical corridor. This suggests that the strength of these methods can be carried over to soft-constrained network training, as long as the waypoint is placed in the right basin (e.g., at corridor intersections). 

\begin{figure*}[t]
\centering
\includegraphics[width=\linewidth]{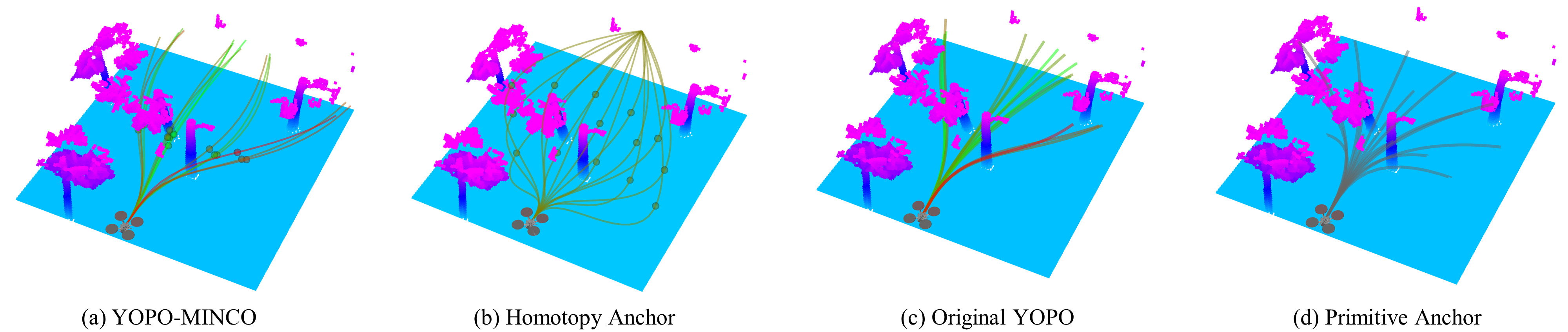}
\caption{Comparison: (a–d) predictions of YOPO-MINCO, the homotopy anchor in this work, the original YOPO, and the original primitive anchor, respectively.}
\label{fig:compare}
\end{figure*}

Without front-end search or safe flight corridors, a learning framework cannot reliably place the intermediate waypoint in the correct basin beforehand. Some methods \cite{learning_optimization} feed the corridor to the network explicitly; we instead exploit the fact that the network predicts in parallel, and lay out a set of preset homotopy anchors---much as YOLO tiles the image with a large number of anchor boxes. We therefore reconsider our strategy through the 2-piece MINCO trajectory representation. The spatio-temporal joint optimization of MINCO~\cite{ref_minco} allows safety to shape the trajectory first, trading extended trajectory duration for smoothness. Meanwhile, we re-express YOPO's primitives as homotopy anchors: each intermediate waypoint anchors a distinct obstacle-avoidance direction, so as to cover different local minima with  higher quality. The endpoint accounts for the goal cost independently, decoupling obstacle avoidance from the goal term. We modify the safety cost to be aggregated by LogSumExp, so that the optimal intermediate-waypoint position converges to the most dangerous instant of the trajectory; the intermediate waypoint therefore becomes a homotopy anchor located at the avoidance bottleneck (i.e., the turning point). A comparison is presented in Fig. \ref{fig:compare}.
Speed also has to be regulated to balance smoothness against safety. Classical reactive schemes do so in velocity space: velocity
obstacles~\cite{ref_vo} exclude velocities that lead to future collisions, and the dynamic
window approach~\cite{ref_dwa} further restricts them to those from which the robot can
brake in time. Learning-based and optimization-based flight systems define speed costs
from environment density or from the projection of the nearest obstacle onto the velocity
direction~\cite{ref_eva}, or compute a required safety margin from the current
speed~\cite{ref_shield}; others learn to adapt the speed or the weights of the classical trajectory optimization to the environment. Time-optimal path parameterization~\cite{ref_topp} decouples speed
planning from path geometry, which is the theoretical basis for regulating speed by
centripetal acceleration. In this spirit, we impose barrier penalties on speed and acceleration, together with a curvature-dependent speed cap whose geometry is detached from the gradient, from which environment-adaptive speed regulation emerges. For example, this avoids high-speed U-turns when the user clicks a new goal.

Finally, many end-to-end driving and navigation systems predict occupancy or other geometric quantities as auxiliary tasks~\cite{ref_cogad} to regularize feature learning, and learned corridors have been used as an interface to a downstream optimizer~\cite{ref_dfto}. Following this line, we predict a safety radius (i.e., safe corridors) for each trajectory, allowing the policy to ``anticipate what will happen after executing a given trajectory." This can be seen as a minimal reduction of a world model for the navigation task: it predicts only safety-relevant quantities, avoiding the prediction of unnecessary depth details.

\section{Method}

\subsection{Overview}
The network consumes a single depth image and the body-frame state
$\bm{o}=[\bm{v},\bm{a},\bm{g}]$ (velocity, acceleration and clipped goal direction), and
predicts for each of the $N$ anchors the raw parameters of one trajectory together
with a scalar score. A differentiable decoder maps the raw parameters to the intermediate
waypoint $\bm{q}_{\text{inner}}$, the terminal state $[\bm{p}_f,\bm{v}_f,\bm{a}_f]$ and the
two segment durations $T_1,T_2$, from which a two-piece MINCO trajectory is reconstructed in
closed form. Intermediate points are predicted as offsets from different anchors (as in YOPO) to ensure multi-modality, while the endpoint is left unconstrained to approach the target. Every candidate is scored by the differentiable costs of safety, smoothness, goal, and feasibility (Sections~\ref{sec:safety}--\ref{sec:goal}), while
the score head is trained by the ranking loss of Section~\ref{sec:rank}.

This partially decouples the competing costs by assigning each to a separate variable: \emph{time for smoothness, the intermediate point for avoidance, and the endpoint for goal reaching}.

\subsection{Two-Piece Spatio-Temporal Representation}
\label{sec:minco}
We enrich the trajectory with additional degrees of freedom through a two-piece, $s=3$
(minimum-jerk) MINCO parameterization~\cite{ref_minco}, so that obstacle avoidance and
goal reaching are taken over by two separate waypoints instead of being shouldered by a
single terminal state: the intermediate waypoint carries the avoidance topology
(Section~\ref{sec:anchor}), while the terminal state carries the goal guidance
(Section~\ref{sec:goal}). Its diffeomorphism
$\mathcal{M}:(\bm{q},\bm{T})\mapsto\bm{c}$ reconstructs the polynomial coefficients
analytically from sparse waypoints and durations, with minimum control effort and
continuity up to order $2s-2$. Smoothness and duration then form one spatio-temporal
objective
\begin{equation}
J = J_{\text{smooth}} + \rho\,T,\qquad
J_{\text{smooth}}=\int_0^{T}\big\lVert\bm{p}^{(3)}(t)\big\rVert^2\,dt,
\label{eq:minco}
\end{equation}
with $T=\sum_k T_k$, both differentiable in $(\bm{q},\bm{T})$.

\subsection{The Intermediate Waypoint as a Homotopy Anchor}
\label{sec:anchor}
Obstacle avoidance is a \emph{multi-homotopy} problem: one may pass an obstacle on either
side but not through it, so the safety cost of a whole trajectory is non-convex with
multiple local minima. Moreover, safety gradients at different positions along the same
trajectory conflict---the first half may need to shift left while the second half needs to
shift right---and cancel on the shared decision variables.

We therefore promote $\bm{q}_{\text{inner}}$ from an extra degree of freedom to a
\emph{homotopy anchor}. Placing it on one side of an obstacle selects a homotopy class and
locks the trajectory into the corresponding feasible basin, inside which the safety cost is
approximately convex and the internal gradient conflict disappears. The ideal location is
the corridor interface, i.e. the \emph{bottleneck} where the ``which way around'' decision
is made; Section~\ref{sec:safety} shows that the focused safety cost drives the anchor
there by itself. This yields the benefit of hard-constrained methods---anchoring a homotopy
class and escaping local minima---without ever forming a corridor. We therefore
lay out a family of homotopy anchors in parallel over the lattice; for the short field of
view of a depth camera two pieces already suffice to express the required detour
topology~\cite{ref_elastic}. We explored predicting spherical corridors with ESDF supervision, similar to detecting safe regions, and then solving MINCO within them. However, this approach did not show clear advantages over directly predicting the trajectory. Moreover, the smoothest and fastest solution can occasionally approach the constraint boundary, where the learned corridors may be less reliable.

\begin{figure}[t]
\centering
\includegraphics[width=\linewidth]{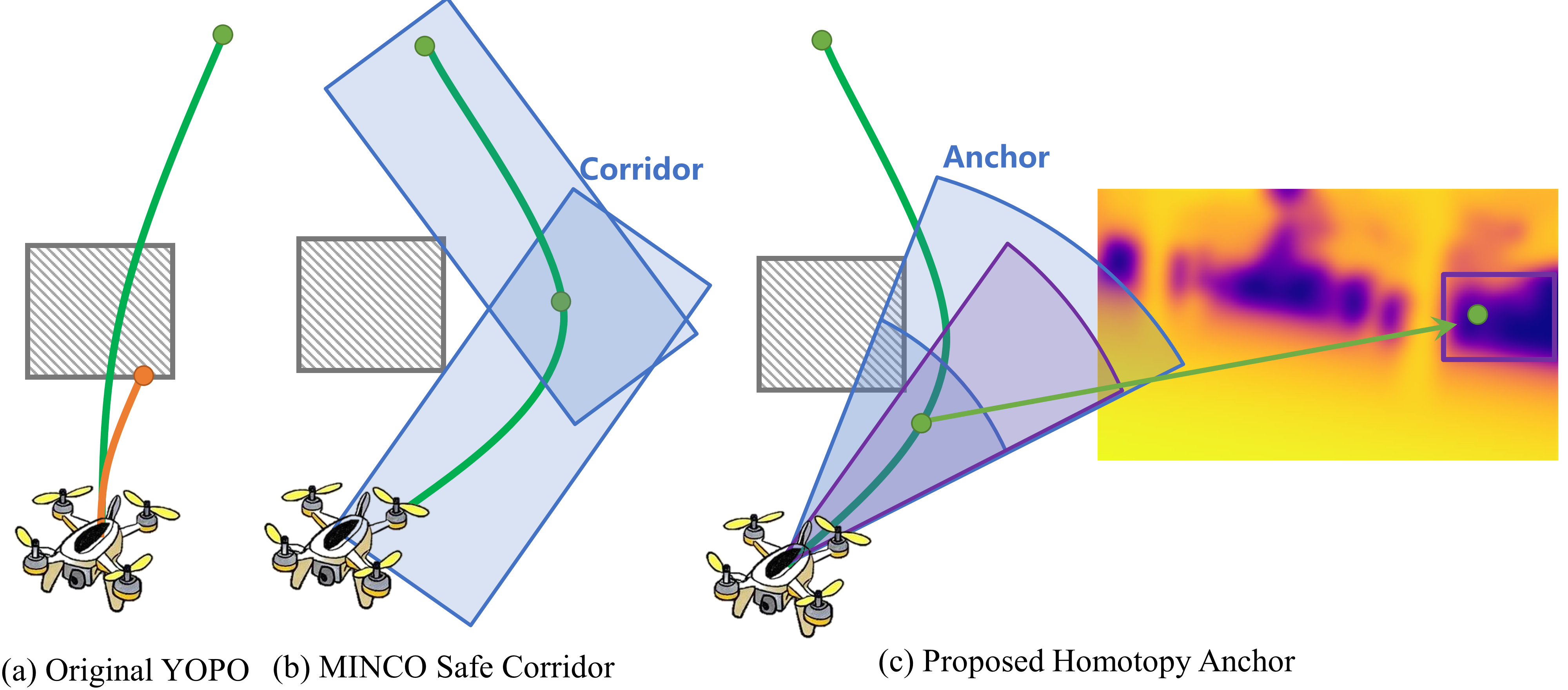}
\caption{In subfigure (c), each frustum defines a fan-shaped optimization corridor for an intermediate point, which can be further constrained by the gradient obtained at its projection on the depth image.}
\label{fig:frustum}
\end{figure}

Following the diffeomorphic idea of MINCO, the network predicts raw parameters in an unbounded space and maps them into a \emph{frustum} (Fig.~\ref{fig:frustum}). This guarantees visibility and a sensible spatial range and, since the frustum of each anchor spans a disjoint region, ensures that the anchors cover mutually exclusive detour topologies. We can project the intermediate waypoint onto the image plane and build a cost from the normalized depth at the projected pixel, so that the first segment conservatively follows an open direction while the second explores a trajectory towards the goal, which may lie behind an obstacle~\cite{ref_safetyassured}.

\subsection{Focused Differentiable Safety Cost}
\label{sec:safety}
Averaging the safety cost along the trajectory is harmful twice over: for a trajectory that
penetrates an obstacle, the samples distributed beyond it contribute gradients that pull it
further in, and a long collision lets the exponential barrier diverge, which also leaves the
score with an unbounded target to predict. We therefore sample densely and aggregate
differently depending on collision. For collision-free trajectories a
LogSumExp soft-maximum focuses the cost on the most dangerous sample,
\begin{equation}
J_{\text{safe}}=\frac{1}{\beta}\log\sum_i\exp\!\big(\beta\,c(\bm{x}_i)\big),
\label{eq:lse}
\end{equation}
with the barrier $c(\bm{x})=\exp\!\big(-(\sdf(\bm{x})-d_0)/r\big)$; $\beta\to\infty$
degenerates to the worst sample and $\beta\to0$ to the mean. For colliding trajectories we
penalize only the last sample \emph{before} the first collision, so the gradient always
points backwards and the trajectory is never pushed to the far side of the obstacle. Every
collision is thus equally bad, independent of penetration depth.

This aggregation is also what places the anchor at the bottleneck. Because the softmax
weights of \eqref{eq:lse} concentrate on the most dangerous instant
$t^\ast=\arg\max_t c(\bm{p}(t))$, both the direction and the magnitude of the gradient are
taken over by that single sample, whereas under an averaged cost the same signal would be
diluted among the harmless ones. The intermediate waypoint, in turn, influences the
trajectory only in the neighborhood of $T_1$, since both ends are pinned by the boundary
conditions and its influence peaks at $t=T_1$. The fastest way to relieve the violation is
therefore to move $T_1$ onto $t^\ast$, where the waypoint has maximal leverage, and then to
push the waypoint along the ESDF gradient; as the trajectory opens up, $t^\ast$ drifts and
$T_1$ follows, so the optimization settles with the waypoint at the bottleneck. % This is the differentiable, anticipatory counterpart of inserting a waypoint constraint wherever a collision is detected~\cite{ref_richter}.

\subsection{Dynamic Feasibility}
\label{sec:feas}
The time cost compresses the duration and raises speed, so speed and acceleration are
clamped by hinge (barrier) penalties on densely sampled states, active only beyond the
dynamic limits and taken at the peak violation,
\begin{equation}
J_{\text{feas}}=\max_i\relu\!\big(\lVert\bm{v}_i\rVert-v_{\max}\big)
             +\max_i\relu\!\big(\lVert\bm{a}_i\rVert-a_{\max}\big).
\end{equation}

\begin{figure}[t]
\centering
\includegraphics[width=0.9\linewidth]{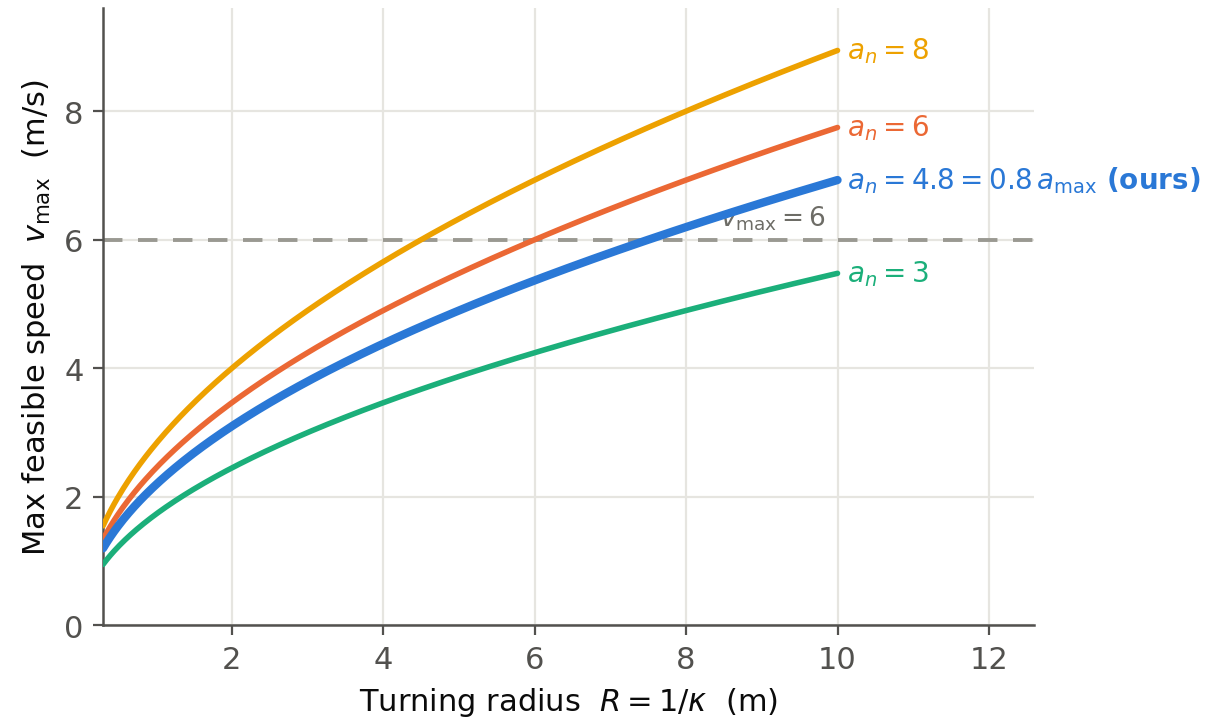}
\caption{Curvature–maximum velocity relationship under different acceleration limits.}
\label{fig:vcap}
\end{figure}

Besides, in a soft-constrained formulation the smoothness term has a
shortcut---it may reshape the trajectory directly, straightening it, instead of adjusting the
duration $T$. A hard-constrained formulation offers no such shortcut, because the corridor
confines the shape and time is the only currency left. Speed must therefore be constrained
explicitly. Cutting the gradient of the smoothness cost with respect to the terminal and
waypoint positions, keeping only its effect on time, is a natural alternative but performed
worse; we instead regulate speed geometrically. The intuitive choice is
to cap the speed by the braking distance afforded by the ESDF clearance, but the ESDF
reports only the \emph{nearest} obstacle, often the ground rather than one along the flight
direction, and a multirotor avoids obstacles by lateral detours rather than by
braking---which introduces a centripetal acceleration constraint. With a normal-acceleration
budget $a_{\text{soft}}=\rho\,a_{\max}$
($\rho<1$ leaves tangential margin) and the path decomposition
$a_n=\kappa v^2$, $\kappa=\lVert\bm{v}\times\bm{a}\rVert/\lVert\bm{v}\rVert^3$, the bound
$a_n\le a_{\text{soft}}$ gives the curvature-dependent speed cap
\begin{equation}
v\le v_{\text{cap}}=\sqrt{a_{\text{soft}}/\kappa},
\label{eq:vcap}
\end{equation}
penalized per sample by $\relu(\lVert\bm{v}_i\rVert-v_{\text{cap},i})$. \emph{Crucially}, the
curvature is detached and the gradient flows only through speed, so ``higher curvature,
lower speed'' never straightens the trajectory the way the smoothness cost does
(Fig.~\ref{fig:vcap}).

When the goal lies to the side, an unconstrained planner flies straight past it. Following
the $L_1$ guidance law, the lateral acceleration required to align the vehicle with the goal
over a look-ahead distance $L_1$ is $a_{\text{cmd}}=2v^2\sin\eta/L_1$. Taking $L_1=d$ (the
goal distance) and $\eta=\theta$ gives the required curvature
$\kappa_{\text{req}}=2\sin\theta/d$ and, under the same budget, the lateral speed limit
\begin{equation}
v_{\text{lim}}=\sqrt{a_{\text{soft}}/\kappa_{\text{req}}}=\sqrt{a_{\text{soft}}\,d/(2\sin\theta)} .
\label{eq:vlim}
\end{equation}
Near-goal deceleration follows the same principle by penalizing the terminal speed once the
goal is close.

Since the required normal acceleration grows with the square of speed, a high-curvature
trajectory necessarily restricts the achievable speed---the rationale of time-optimal path
parameterization~\cite{ref_topp} for decoupling speed planning from path geometry.
Environment-adaptive speed then emerges from \eqref{eq:vcap}: the more cluttered the scene,
the larger the curvature needed to avoid obstacles and maintain the homotopy class, hence
the lower the speed bound, and MINCO extends the corresponding duration accordingly; in open
space the trajectory straightens, the bound is released and the duration shortens.

\subsection{Ranking-Based Candidate Evaluation}
\label{sec:rank}
We initially regressed the trajectory cost as the score. Only the safety cost must be
predicted, since the smoothness and goal terms can be evaluated in closed form at runtime,
yet even that is hard: obstacles may be occluded, the cost grows very fast under
penetration, and the collision duration is difficult to estimate. More importantly, a small
score error can badly reorder the candidates. We therefore supervise the score head by ranking \cite{rank}, with a
one-hot target marking the cheapest \emph{collision-free} candidate,
\begin{equation}
J_{\text{rank}}=\mathrm{CE}\!\Big(\bm{s},\;
\arg\min_n\big[J_n+\Lambda\cdot\mathbb{1}(\text{collide}_n)\big]\Big),
\end{equation}
where the large offset $\Lambda$ ranks colliding candidates behind all feasible ones and
degenerates to the least-bad one when all collide.

The generation loss acts on the total cost of each candidate. Since training states are
sampled at random, many fall into dangerous regions. This is not necessarily a drawback, as it exposes the model to diverse situations to handle arbitrary observations. As the safety cost grows
exponentially while the smoothness cost grows with a high power of speed, those samples would
dominate the gradient. We therefore normalize the $N$ candidates of each image by their mean,
\begin{equation}
\tilde J_n=J_n\Big/\Big(\tfrac1N\textstyle\sum_m J_m\Big),
\label{eq:norm}
\end{equation}
so that hazardous and benign samples contribute equally regardless of their absolute cost
magnitude, and weight the result by the \emph{detached} score,
$J_{\text{gen}}=\frac{1}{BN}\sum_n\sg(s_n)\,\tilde J_n$, concentrating optimization on the
candidates the network considers good.

\subsection{Time-Scale Invariance}
\label{sec:scale}
Being piecewise polynomial, the representation is time-scale invariant, so one policy covers
a range of cruise speeds. For a piece $\bm{p}(t)=\sum_k\bm{c}_kt^k$ and the scaling
$t\mapsto t/\alpha$,
\begin{equation}
\tilde{\bm{p}}(t)=\bm{p}(t/\alpha)=\sum_k\bm{c}_k\alpha^{-k}t^k,\quad t\in[0,\alpha T],
\end{equation}
the spatial path is identical and
$\tilde{\bm{p}}^{(n)}(t)=\alpha^{-n}\bm{p}^{(n)}(t/\alpha)$, so velocity, acceleration and
jerk scale by $\alpha^{-1},\alpha^{-2},\alpha^{-3}$. The minimum-jerk energy obeys
$\int_0^{\alpha T}\lVert\tilde{\bm{p}}^{(3)}\rVert^2dt=\alpha^{-5}\int_0^{T}\lVert\bm{p}^{(3)}\rVert^2dt$,
a monotone rescaling for a fixed geometry, so the optimum remains optimal within the same
geometric family. We train at $v^{\text{train}}_{\max}$ and deploy at $v$ with
$\alpha=v^{\text{train}}_{\max}/v$: the predicted geometry and topology are unchanged and
only the time axis is scaled.

\subsection{Safe-Corridor Radius Prediction}
\label{sec:corridor}
\begin{figure}[t]
\centering
\includegraphics[width=0.65\linewidth]{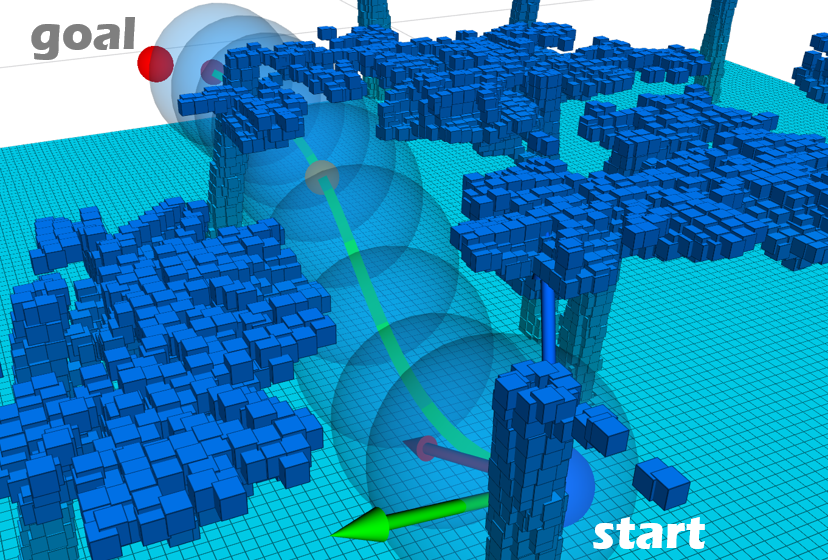}
\caption{Predicted safe-corridor radii along a candidate trajectory.}
\label{fig:corridor}
\end{figure}

Each candidate additionally predicts the minimum clearance at $K$ instants along its
trajectory, letting the network anticipate what will happen once its own prediction is
executed (Fig.~\ref{fig:corridor}). It can also serve as a corridor for a downstream
refinement and as an auxiliary task supervising feature learning
towards correct geometry, as in methods predicting occupancy~\cite{ref_cogad}. The
distance is first warped by $y=1-\exp(-d/\lambda)$, concentrating precision near obstacles,
and supervised by an asymmetric Laplace negative log-likelihood
\begin{equation}
\mathcal{L}=w\,\frac{|y-\mu|}{b}+\log b ,
\end{equation}
in which over-estimation is weighted more heavily (a corridor must contain no obstacle),
forcing $\mu$ to a conservative quantile. The scale $b$ lets occluded or out-of-view samples
release pressure by inflating $b$ instead of pulling $\mu$ towards the dataset mean. The two
terms are the negative log-likelihood of $p(y\mid\mu,b)=\frac{1}{2b}\exp(-|y-\mu|/b)$, so the
output is not a point estimate but the statement ``the truth is about $\mu$, within $\pm b$'';
setting $\partial\mathcal{L}/\partial b=0$ gives $b^\star=w\,|e|$, the expected absolute error
at that point.

However, this auxiliary task brings only marginal value. The
trajectory head already receives exact per-sample geometric gradients through the
differentiable ESDF, so the corridor task adds almost no information. The negative result is
itself good news: the learned features are already geometrically correct and need no
additional supervision. We nevertheless keep the design as a safety fallback, since it comes
at no cost.

% The field is seeded at a \emph{virtual goal} $15$~m along the goal direction while the
% progress reward is truncated at the \emph{nominal} distance ($10$~m). The two play different
% roles: the $15$~m field sets the \emph{direction} of the pull, since the level sets near the
% nominal goal still flow towards the virtual one, so the trajectory anticipates obstacles
% between $10$ and $15$~m; the truncation sets where the pull \emph{ends}, clamping the cost to
% zero at and beyond the nominal distance so the endpoint is drawn to $10$~m without rewarding
% or penalizing overshoot.

\section{Others}

\subsection{Geodesic Goal Guidance Field}
\label{sec:goal}
Although the intermediate point and the endpoint handle safety and goal reaching, respectively, the safety and goal costs can still compete around large obstacles.   An overly high ESDF-induced cost forms high-cost saddles in the gaps between obstacles, sealing them off, while the Euclidean distance to the goal induces trajectories that cut into the obstacles. The lattice supplies a multi-modal prior in cluttered scenes, but large
obstacles call for stronger look-ahead guidance. A guidance path is single-modal and
degenerates the cost into a distance to one labeled path; the Voronoi potential of hybrid
A$^\ast$~\cite{ref_hybridastar} keeps narrow passages traversable, avoiding the saddle points
that the ESDF leaves between obstacles.

\begin{figure}[t]
\centering
\includegraphics[width=\linewidth]{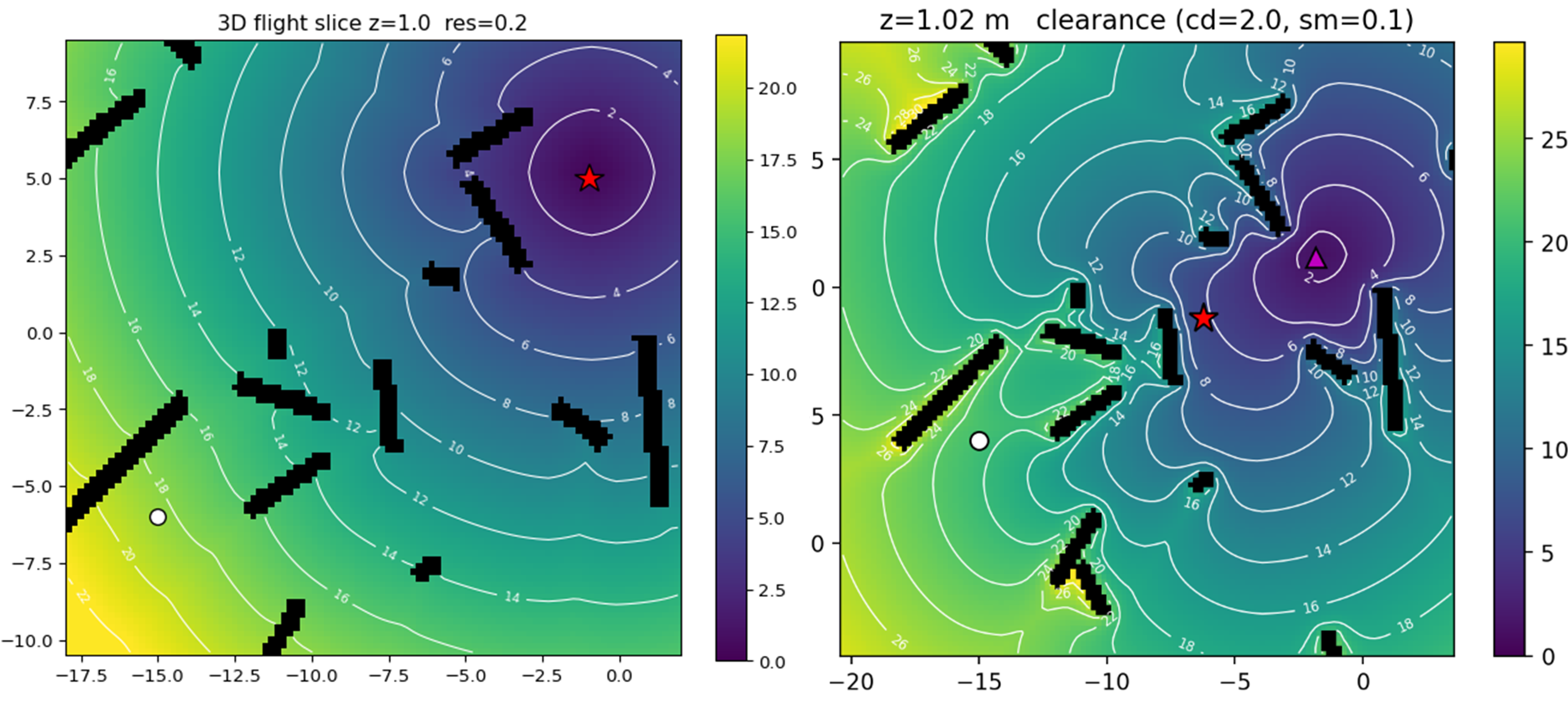}
\caption{Cost-to-go field visualization, with the right one coupled with the ESDF.}
\label{fig:field}
\end{figure}

We instead replace the Euclidean goal cost by a Dijkstra-type \emph{geodesic cost-to-go
field} $V$, obtained by solving the eikonal equation
$\lVert\nabla V(\bm{x})\rVert=1/f(\bm{x})$ with the Fast Marching Method~\cite{ref_fmm}, the
continuous counterpart of Dijkstra's algorithm, with the traversal speed $f$ lowered near
obstacles (Fig.~\ref{fig:field}). $V$ is the value function of the corresponding minimum-time problem
over the whole free space: clearance is embedded in the geodesic distance itself, and
descending $-\nabla V$ traces the minimum-cost path to the goal. Unlike a guidance path, it
gives every candidate a valid descent direction whatever its homotopy class, instead of
collapsing them onto one path; and since the geodesics bend around obstacles, the guidance
is far-sighted and does not conflict with the safety cost.

\subsection{Temporal Consistency}
Temporal consistency is not explicitly considered: to keep the training method simple and elegant, the network decides
independently at every frame. This is not a serious issue, since many classical methods also follow a front-end-search / back-end-optimization scheme without considering the previous decision, and the problem can be mitigated indirectly in various ways. For example, (1) enforcing continuity of the initial state together with a high-order trajectory representation avoids the discontinuities that arise from directly predicting control commands; (2) among the multiple optimal candidates, selecting the one closest to the past trajectory, similar to \cite{SR}; and (3) more fundamentally, with larger-scale data the network can learn to produce consistent decisions in similar scenarios, as suggested by self-supervised vision models: DINO~\cite{ref_dino} is trained on still images with no temporal supervision, yet its per-frame features are stable enough across consecutive frames.

\subsection{Implicit Positional Encoding}
Convolutions are translation invariant and treat all feature-map locations alike, which is
at odds with planning, where different image regions correspond to different topologies. We
therefore express the observation and the prediction of every cell in its own primitive
frame, so the task becomes equivalent across cells and all $N$ primitives share one head---
removing the ambiguity of ``the same output meaning different topologies at different
locations'' and accelerating convergence, since one forward pass is approximately $N$
supervisions of the planning head. This plays the role of positional encoding in
transformers and can equivalently be replaced by concatenating the pitch and yaw angles of
each primitive to the observation.

Removing the transformation, however, does not degrade performance noticeably, because zero
padding already leaks absolute position: boundary padding makes the receptive-field
statistics of different cells systematically different, which the network exploits as a free
positional encoding. Islam \textit{et al.}~\cite{ref_padding} show that freezing an ordinary
ResNet without positional embedding and regressing image coordinates succeeds whenever zero
padding is present.

\subsection{Sim-to-Real Generalization}
\begin{figure}[t]
\centering
\includegraphics[width=\linewidth]{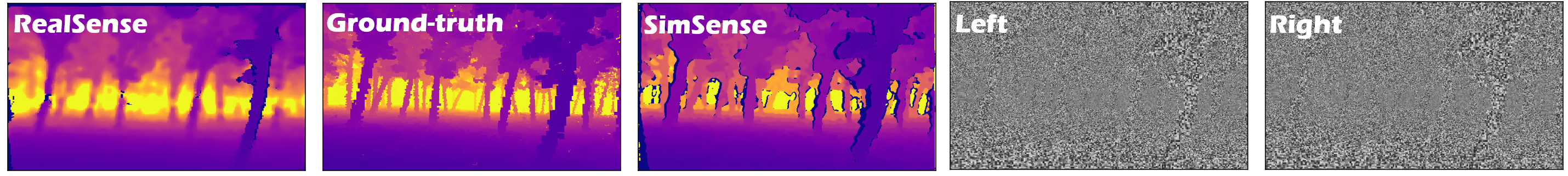}
\vspace{-0.6cm}
\caption{Comparison of RealSense, ground-truth, and stereo depth. Right subfigures: rendered speckle texture for stereo matching.}
\label{fig:stereo}
\end{figure}

\begin{figure}[t]
\centering
\includegraphics[width=\linewidth]{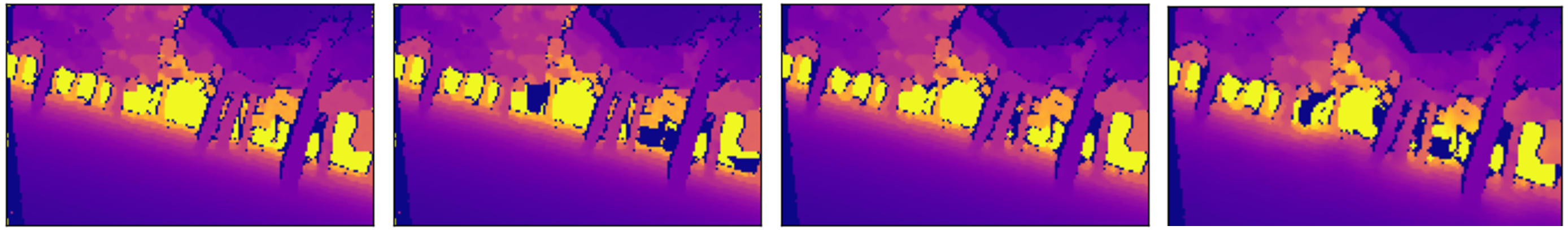}
\vspace{-0.5cm}
\caption{Depth augmentation with random noise and missing regions.}
\label{fig:aug}
\end{figure}

To evaluate sim-to-real generalization, a real-world dataset is collected with a
RealSense camera and a LiDAR to serve as the validation set. We augment our CUDA simulator with structured-light stereo
imaging: a speckle texture is rendered in the world and RealSense-style depth is computed by
SimSense~\cite{ref_simsense} (stereo/pseudo-infrared with SGM), Fig.~\ref{fig:stereo}, and
we experiment with a variety of augmentation strategies, including erosion, holes, noise and
occlusion (Fig. \ref{fig:aug}).
However, none of the strategies we tried---stereo matching, lower resolution, and depth
augmentation---brought a noticeable gain in generalization over simply training on
ground-truth depth.

\begin{figure}[t]
\centering
\includegraphics[width=\linewidth]{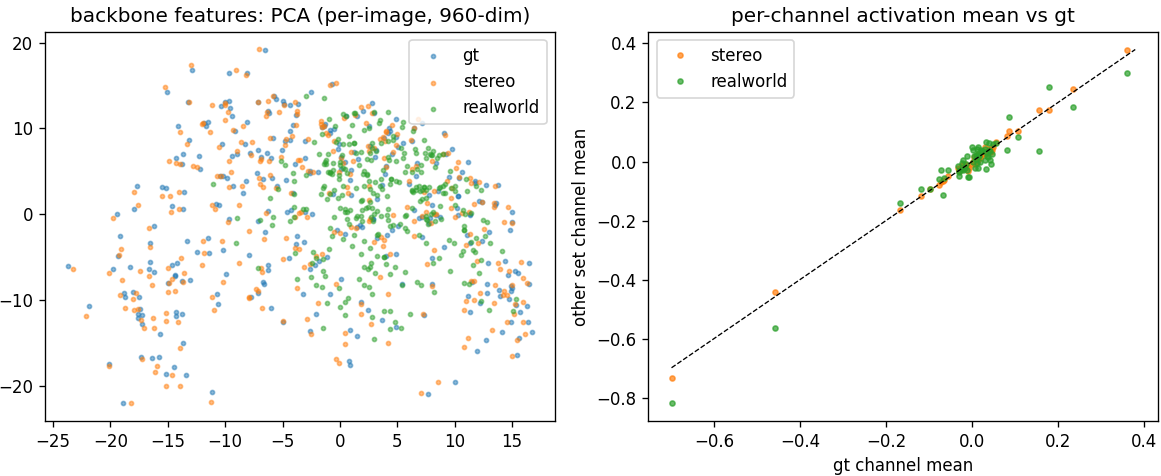}
\caption{t-SNE of features from simulated ground-truth depth / stereo depth and RealSense depth images.}
\label{fig:feature}
\end{figure}

Besides, to determine whether the unsafety is caused by network adaptation or by depth inaccuracies themselves, we train the network with ground-truth depth in simulation and evaluate it on the real-world dataset using two collision metrics: collisions against the ground-truth environmental point cloud and collisions against obstacles projected from the RealSense depth image. The collision rate evaluated from the projected depth image is much lower than that evaluated against the ground-truth point cloud.
This indicates that the network effectively avoids obstacles captured in the depth images and is robust to the image style. But perception is not always
accurate in the real world. Training on ground truth and evaluating with a RealSense D455, most collisions occur at
\emph{long range}, where the sensor reports obstacles as farther than they are. Besides, an effective
remedy for sim-to-real is contrastive learning \cite{gap}, which forces the network to extract
consistent features from images of different styles. In our method, however, a t-SNE
comparison (Fig.~\ref{fig:feature}) shows that the features our network extracts from
ground-truth depth and from stereo depth are already very close: the gap
stems from scene statistics and depth error rather than from the stereo modality. Furthermore, compared with no inpainting during training and testing, inpainting invalid depth values can reduce the sim-to-real gap.

The sim-to-real generalization is important. However, the test results suggest that the network appears relatively robust to image style and noise, with the main issue arising from perception errors that increase with distance. Therefore, we do not
overfit the error model of specific sensor: this is a perception problem, and the planner should
avoid what is \emph{perceived}. 

\section{Conclusion}
We revisited the representation, the cost design and the training signal of one-stage
reactive planners. We adopt a two-piece MINCO trajectory representation, which lets the
intermediate waypoint take over the avoidance topology, and a LogSumExp safety cost, which
drives that waypoint to the bottleneck. A curvature-dependent speed cap with
detached geometry then makes speed adapt to the environment, a geodesic cost-to-go field
gives every homotopy class a correct descent direction, and a ranking loss makes selection
robust. The negative results are informative too: corridor prediction, explicit positional
encoding and depth augmentation all brought little, suggesting that the dense differentiable
supervision already yields geometrically correct features and that the remaining collisions
are a perception rather than a planning problem. This report is intended primarily to document our open-source branch and attempts. A full experimental evaluation is left for future work.

\bibliographystyle{IEEEtran}
\bibliography{refs}

% Generated by IEEEtran.bst, version: 1.14 (2015/08/26)
\begin{thebibliography}{10}
\providecommand{\url}[1]{#1}
\csname url@samestyle\endcsname
\providecommand{\newblock}{\relax}
\providecommand{\bibinfo}[2]{#2}
\providecommand{\BIBentrySTDinterwordspacing}{\spaceskip=0pt\relax}
\providecommand{\BIBentryALTinterwordstretchfactor}{4}
\providecommand{\BIBentryALTinterwordspacing}{\spaceskip=\fontdimen2\font plus
\BIBentryALTinterwordstretchfactor\fontdimen3\font minus
  \fontdimen4\font\relax}
\providecommand{\BIBforeignlanguage}[2]{{%
\expandafter\ifx\csname l@#1\endcsname\relax
\typeout{** WARNING: IEEEtran.bst: No hyphenation pattern has been}%
\typeout{** loaded for the language `#1'. Using the pattern for}%
\typeout{** the default language instead.}%
\else
\language=\csname l@#1\endcsname
\fi
#2}}
\providecommand{\BIBdecl}{\relax}
\BIBdecl

\bibitem{ref_yopo}
J.~Lu, X.~Zhang, H.~Shen, L.~Xu, and B.~Tian, ``You only plan once: A
  learning-based one-stage planner with guidance learning,'' \emph{IEEE
  Robotics and Automation Letters}, vol.~9, no.~7, pp. 6083--6090, 2024.

\bibitem{fast}
B.~Zhou, F.~Gao, L.~Wang, C.~Liu, and S.~Shen, ``Robust and efficient quadrotor
  trajectory generation for fast autonomous flight,'' \emph{IEEE Robotics and
  Automation Letters}, vol.~4, no.~4, pp. 3529--3536, 2019.

\bibitem{SR}
A.~Loquercio, E.~Kaufmann, R.~Ranftl, M.~M{\"u}ller, V.~Koltun, and
  D.~Scaramuzza, ``Learning high-speed flight in the wild,'' \emph{Science
  Robotics}, vol.~6, no.~59, p. eabg5810, 2021.

\bibitem{Deep-panther}
J.~Tordesillas and J.~P. How, ``Deep-panther: Learning-based perception-aware
  trajectory planner in dynamic environments,'' \emph{IEEE Robotics and
  Automation Letters}, vol.~8, no.~3, pp. 1399--1406, 2023.

\bibitem{back}
Y.~Zhang, Y.~Hu, Y.~Song, D.~Zou, and W.~Lin, ``Learning vision-based agile
  flight via differentiable physics,'' \emph{Nature Machine Intelligence},
  vol.~7, no.~6, pp. 954--966, 2025.

\bibitem{racing}
Y.~Song, A.~Romero, M.~M{\"u}ller, V.~Koltun, and D.~Scaramuzza, ``Reaching the
  limit in autonomous racing: Optimal control versus reinforcement learning,''
  \emph{Science Robotics}, vol.~8, no.~82, p. eadg1462, 2023.

\bibitem{ref_bezier}
F.~Gao, W.~Wu, Y.~Lin, and S.~Shen, ``Online safe trajectory generation for
  quadrotors using fast marching method and bernstein basis polynomial,'' in
  \emph{2018 IEEE international conference on robotics and automation
  (ICRA)}.\hskip 1em plus 0.5em minus 0.4em\relax IEEE, 2018, pp. 344--351.

\bibitem{ref_minco}
Z.~Wang, X.~Zhou, C.~Xu, and F.~Gao, ``Geometrically constrained trajectory
  optimization for multicopters,'' \emph{IEEE Transactions on Robotics},
  vol.~38, no.~5, pp. 3259--3278, 2022.

\bibitem{ref_bubble}
Y.~Ren, F.~Zhu, W.~Liu, Z.~Wang, Y.~Lin, F.~Gao, and F.~Zhang, ``Bubble
  planner: Planning high-speed smooth quadrotor trajectories using receding
  corridors,'' in \emph{2022 IEEE/RSJ International Conference on Intelligent
  Robots and Systems (IROS)}.\hskip 1em plus 0.5em minus 0.4em\relax IEEE,
  2022, pp. 6332--6339.

\bibitem{learning_optimization}
Y.~Wu, X.~Sun, I.~Spasojevic, and V.~Kumar, ``Deep learning for optimization of
  trajectories for quadrotors,'' \emph{IEEE Robotics and Automation Letters},
  vol.~9, no.~3, pp. 2479--2486, 2024.

\bibitem{ref_vo}
P.~Fiorini and Z.~Shiller, ``Motion planning in dynamic environments using
  velocity obstacles,'' \emph{The international journal of robotics research},
  vol.~17, no.~7, pp. 760--772, 1998.

\bibitem{ref_dwa}
D.~Fox, W.~Burgard, and S.~Thrun, ``The dynamic window approach to collision
  avoidance,'' \emph{IEEE robotics \& automation magazine}, vol.~4, no.~1, pp.
  23--33, 1997.

\bibitem{ref_eva}
L.~Quan, Z.~Zhang, X.~Zhong, C.~Xu, and F.~Gao, ``Eva-planner: Environmental
  adaptive quadrotor planning,'' in \emph{2021 IEEE International Conference on
  Robotics and Automation (ICRA)}.\hskip 1em plus 0.5em minus 0.4em\relax IEEE,
  2021, pp. 398--404.

\bibitem{ref_shield}
J.~Zhang, C.~Lei, C.~Dai, K.~Hoi, L.~Wang, Z.~Han, and F.~Gao, ``High-speed
  vision-based flight in clutter with safety-shielded reinforcement learning,''
  \emph{IEEE Robotics and Automation Letters}, 2026.

\bibitem{ref_topp}
H.~Pham and Q.-C. Pham, ``A new approach to time-optimal path parameterization
  based on reachability analysis,'' \emph{IEEE Transactions on Robotics},
  vol.~34, no.~3, pp. 645--659, 2018.

\bibitem{ref_cogad}
Z.~Wang, J.~Teng, C.~Xiang, K.~Chen, X.~Pan, L.~Deng, and W.~Gu, ``Cogad:
  Cognitive-hierarchy guided end-to-end autonomous driving,'' in
  \emph{Proceedings of the IEEE/CVF Conference on Computer Vision and Pattern
  Recognition}, 2026, pp. 967--977.

\bibitem{ref_dfto}
Z.~Han, L.~Xu, L.~Pei, and F.~Gao, ``Dynamically feasible trajectory generation
  with optimization-embedded networks for autonomous flight,'' \emph{IEEE
  Robotics and Automation Letters}, 2025.

\bibitem{ref_elastic}
J.~Ji, N.~Pan, C.~Xu, and F.~Gao, ``Elastic tracker: A spatio-temporal
  trajectory planner for flexible aerial tracking,'' in \emph{2022
  International Conference on Robotics and Automation (ICRA)}.\hskip 1em plus
  0.5em minus 0.4em\relax IEEE, 2022, pp. 47--53.

\bibitem{ref_safetyassured}
Y.~Ren, F.~Zhu, G.~Lu, Y.~Cai, L.~Yin, F.~Kong, J.~Lin, N.~Chen, and F.~Zhang,
  ``Safety-assured high-speed navigation for mavs,'' \emph{Science Robotics},
  vol.~10, no.~98, p. eado6187, 2025.

\bibitem{ref_richter}
C.~Richter, A.~Bry, and N.~Roy, ``Polynomial trajectory planning for aggressive
  quadrotor flight in dense indoor environments,'' in \emph{Robotics Research:
  The 16th International Symposium ISRR}.\hskip 1em plus 0.5em minus
  0.4em\relax Springer, 2016, pp. 649--666.

\bibitem{rank}
Z.~Cao, T.~Qin, T.-Y. Liu, M.-F. Tsai, and H.~Li, ``Learning to rank: from
  pairwise approach to listwise approach,'' in \emph{Proceedings of the 24th
  international conference on Machine learning}, 2007, pp. 129--136.

\bibitem{ref_hybridastar}
D.~Dolgov, S.~Thrun, M.~Montemerlo, and J.~Diebel, ``Practical search
  techniques in path planning for autonomous driving,'' \emph{ann arbor}, vol.
  1001, no. 48105, pp. 18--80, 2008.

\bibitem{ref_fmm}
J.~A. Sethian, ``A fast marching level set method for monotonically advancing
  fronts.'' \emph{proceedings of the National Academy of Sciences}, vol.~93,
  no.~4, pp. 1591--1595, 1996.

\bibitem{ref_dino}
M.~Caron, H.~Touvron, I.~Misra, H.~J{\'e}gou, J.~Mairal, P.~Bojanowski, and
  A.~Joulin, ``Emerging properties in self-supervised vision transformers,'' in
  \emph{2021 IEEE/CVF international conference on computer vision
  (ICCV)}.\hskip 1em plus 0.5em minus 0.4em\relax IEEE, 2021, pp. 9630--9640.

\bibitem{ref_padding}
M.~A. Islam, S.~Jia, and N.~D. Bruce, ``How much position information do
  convolutional neural networks encode?'' \emph{arXiv preprint
  arXiv:2001.08248}, 2020.

\bibitem{ref_simsense}
\BIBentryALTinterwordspacing
L.~Ang, ``Simsense: A real-time depth sensor simulator,'' 2026. [Online].
  Available: \url{https://github.com/angli66/simsense}
\BIBentrySTDinterwordspacing

\bibitem{gap}
H.~Yu, C.~De~Wagter, and G.~C.~E. de~Croon, ``Depth transfer: Learning to see
  like a simulator for real-world drone navigation,'' \emph{IEEE Robotics and
  Automation Letters}, 2025.

\end{thebibliography}

\end{document}